\documentclass[10pt,twocolumn,letterpaper]{article}
\usepackage{cvpr}
\usepackage{times}
\usepackage{epsfig}
\usepackage{graphicx}
\usepackage{amsmath}
\usepackage{amssymb}
\graphicspath{{./figures/}}

\usepackage{multirow}
\usepackage{array}
\newcolumntype{P}[1]{>{\centering\arraybackslash}p{#1}}
\newcolumntype{M}[1]{>{\centering\arraybackslash}m{#1}}

\usepackage[pagebackref=true,breaklinks=true,letterpaper=true,colorlinks,bookmarks=false]{hyperref}

\cvprfinalcopy % *** Uncomment this line for the final submission

\def\cvprPaperID{111} % *** Enter the CVPR Paper ID here

\ifcvprfinal\fi
\begin{document}

%%%%%%%%% TITLE
\title{Object Detection from Video Tubelets with Convolutional Neural Networks}

\author{Kai Kang\ \ \ \ Wanli Ouyang\ \ \ \ Hongsheng Li\ \ \ \ Xiaogang Wang\\
Department of Electronic Engineering, The Chinese University of Hong Kong\\
{\tt\small \{kkang,wlouyang,hsli,xgwang\}@ee.cuhk.edu.hk}
}

\maketitle
\thispagestyle{empty}

%%%%%%%%% ABSTRACT
\begin{abstract}
Deep Convolution Neural Networks (CNNs) have shown impressive performance in various vision tasks such as image classification, object detection and semantic segmentation. For object detection, particularly in still images, the performance has been significantly increased last year thanks to powerful deep networks (\eg GoogleNet) and detection frameworks (\eg Regions with CNN features (R-CNN)). The lately introduced ImageNet \cite{deng2009imagenet} task on object detection from video (VID) brings the object detection task into the video domain, in which objects' locations at each frame are required to be annotated with bounding boxes. In this work, we introduce a complete framework for the VID task based on still-image object detection and general object tracking. Their relations and contributions in the VID task are thoroughly studied and evaluated. In addition, a temporal convolution network is proposed to incorporate temporal information to regularize the detection results and shows its effectiveness for the task. Code is available at \url{https://github.com/myfavouritekk/vdetlib}.
\end{abstract}

\section{Introduction} % (fold)
\label{sec:intro}
Deep learning has been widely applied to various computer vision tasks such as image classification \cite{krizhevsky2012imagenet,vgg,googlenet}, object detection \cite{girshick2014rich,girshick2015fast,girshick2015faster,googlenet,ouyang2015deepid,ouyang2016factors}, semantic segmentation \cite{long2015fully,kang2014fully}, human pose estimation \cite{toshev2014deeppose,tompson2015efficient,yang2016end,chu2016structure}, \etc.
Over the past few years, the performance of object detection in ImageNet and PASCAL VOC  has been increased by a significant margin with the success of deep Convolutional Neural Networks (CNN).
State-of-the-art methods for object detection train  CNNs to classify region proposals into background or one of the object classes.
% This type of region-proposal-based pipelines also shows strong performance on other tasks such as object segmentation \cite{dai2015boxsup}.
However, these methods focus on detecting objects in still images. The lately introduced ImageNet challenge on object detection from video brings up a new question on how to solve the object detection problem for videos effectively and robustly.
At each frame of a video, the algorithm is required to annotate bounding boxes and confidence scores on objects of each class.
Although there have been methods on detecting objects in videos, they mainly focus on detecting one specific class of objects, such as pedestrians \cite{tian2014pedestrian}, cars \cite{li2014integrating}, or humans with actions \cite{jain2014action,gkioxari2014finding}.
The ImageNet challenge defines a new problem on detecting general objects in videos, which is worth studying.
Similar to object detection in still images being able to assist tasks including image classification \cite{vgg}, object segmentation \cite{dai2015boxsup}, and image captioning \cite{fang2014captions}, accurately detecting objects in videos could possibly boost the performance of video classification, video captioning and related surveillance applications.
By locating objects in the videos, the semantic meaning of a video could also be described more clearly, which results in more robust performance for video-based tasks.

\begin{figure}[t]
    \centering
    \includegraphics[width=0.9\linewidth]{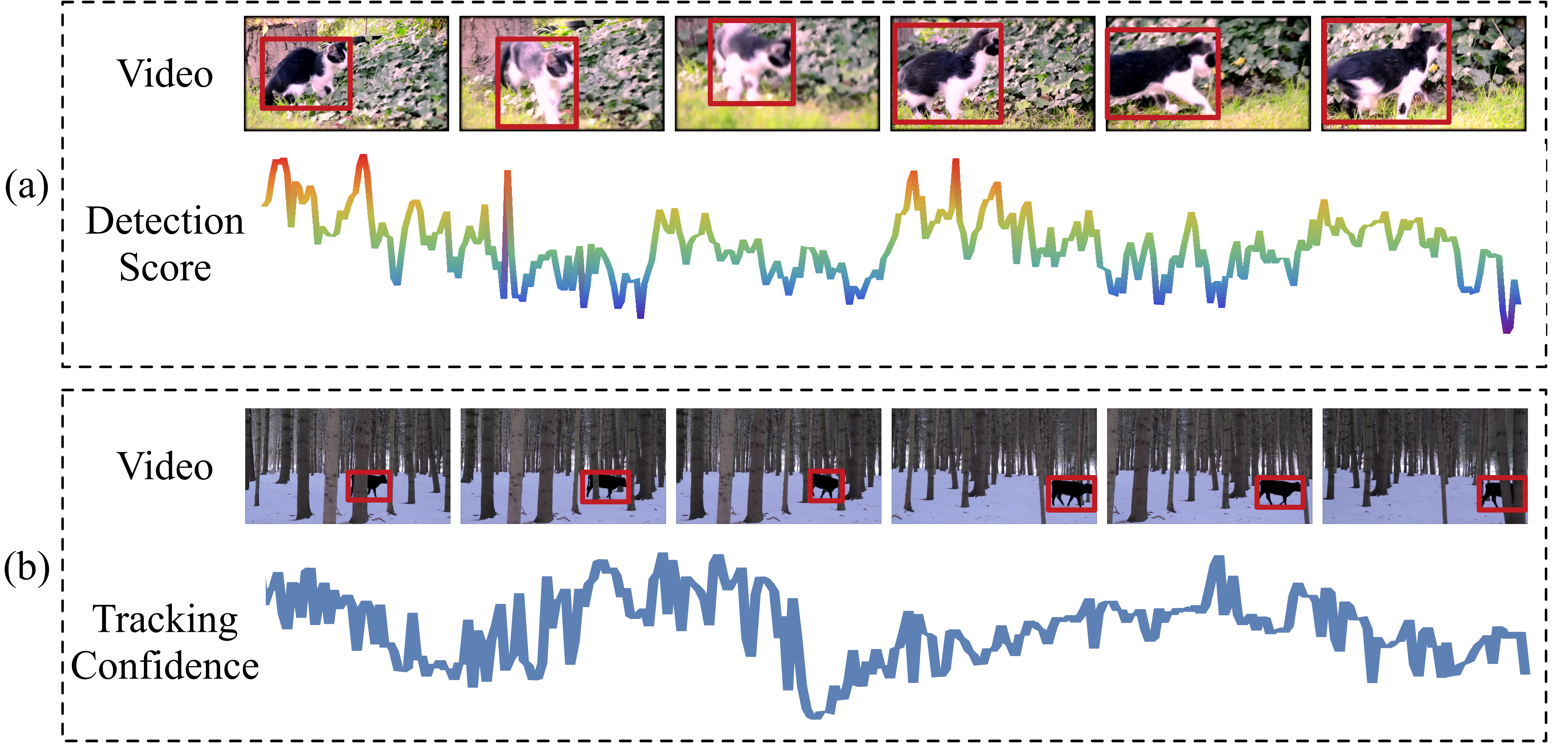}
    \caption{Challenges in object detection from video. Red boxes are ground truth annotations. (a) Still-image object detection methods have large temporal fluctuations across frames even on ground truth bounding boxes. The fluctuations may result from motion blur, video defocus, part occlusion and bad pose.
    % In addition, because the detectors on still-images do not have temporal regulation during training, they are not temporal invariant. Even small changes of normal pose may cause large variations in detection scores.
    Information of boxes of the same object on adjacent frames need to be utilized for object detection in video. (b) Tracking is able to relate boxes of the same object. However, due to occlusions, appearance changes and pose variations, the tracked boxes may drift to non-target objects. Object detectors should be incorporated into tracking algorithm to constantly start new tracks when drifting occurs.}
    \label{fig:challenges}
\end{figure}

Existing methods for general object detection cannot be applied to solve this problem effectively. Their performance may suffer from large appearance changes of objects in videos.
For instance in Figure~\ref{fig:challenges} (a), if a cat faces the camera at first and then turns back. Its back image cannot be effectively recognized as a cat because it contains little texture information and is not likely to be included in training samples.
The correct recognition result needs to be inferred from information in previous and future frames, because the appearance of an object in video is highly correlated.
Since an object's location might change in the video, the location correspondences across the video should be recovered such that the correlated image patches could be well aligned into trajectories for extracting the temporal information.
Besides, temporal consistency of recognition results should be regularized (Figure~\ref{fig:challenges} (a)). The detection scores of a bounding-box tubelet representing an object should not change dramatically across the video.

Such requirements motivate us to incorporate object tracking into our detection framework. Deep CNNs have shown impressive performance on object tracking \cite{wang2015visual,nam2015learning}, which outperform previous methods by a large margin.
The large number of tracking-by-detection methods \cite{andriluka2008people,andriyenko2012discrete,bae2014robust,possegger2014occlusion} for multi-pedestrian tracking have shown that temporal information could be utilized to regularize the detection results.
However, directly utilizing object tracking cannot effectively solve the VID problem either (Figure~\ref{fig:challenges} (b)).
In our experiments, we have noticed that directly using still-image object detectors on object tracking results has only $37.4\%$ mean average precision (mean AP) compared to $45.3\%$ on object proposals. The performance difference results from detectors' sensitivity to location changes and the box mismatch between tracks and object proposals. To solve this problem, we proposed a tubelet box perturbation and max-pooling process to increase the performance from $37.4\%$ to $45.2\%$, which is comparable to the performance of image object proposal with only $1/38$ the number of boxes.
% Even if an initial bounding box for an object is correct, the tracked bounding boxes in the consecutive frames might gradually drift away. This is because tracking methods model the appearance of the tracked objects with information mostly within the video. Such model tracking the target object might gradually shift to track nearby objects.
% It is therefore worthy of investigating how to combine the deep CNNs for both object detection and object tracking to effectively solve the problem of object detection in videos.

In this work, we propose a multi-stage framework based on deep CNN detection and tracking for object detection in videos.
The framework consists of two main modules:
1) a tubelet proposal module that combines object detection and object tracking for tubelet object proposal;
2) a tubelet classification and re-scoring module that performs spatial max-pooling for robust box scoring and temporal convolution for incorporating temporal consistency.
%on samples around tubelet boxes to reduce tracking failures, classifies tubelet proposals for each specific class and re-scores the tubelets using a specially designed temporal CNN to incorporate temporal information and consistency.
Object detection and tracking work closely in our framework. On one hand, object detection produces high-confidence anchors to initiate tracking and reduces tracking failure by spatial max-pooling. On the other hand, tracking also generates new proposals for object detection and the tracked boxes act as anchors to aggregate existing detections.

The contribution of this paper is three-fold.
% ##contributions
% - Propose a complete framework for object detection in videos
% - Study each component in the framework and analysis their importance
% - Propose a temporal convolution network for incorporating temporal features into video detection 
1) A complete multi-stage framework is proposed for object detection in videos.
2) The relation between still-image object detection and object tracking, and their influences on object detection from video are studied in details.
3) A special temporal convolutional neural network is proposed to incorporate temporal information into object detection from video.

\section{Related Works} % (fold)
\label{sec:related_works}
State-of-the-art methods for detecting objects of general classes are mainly based on deep CNNs. Girshick \etal \cite{girshick2014rich} proposed a multi-stage pipeline called Regions with Convolutional Neural Networks
(R-CNN) for training deep CNN to classify region proposals for object detection. It decomposes the detection problem into several stages including bounding-box proposal, CNN pre-training, CNN fine-tuning, SVM training, and bounding box regression. Such framework has shown good performance and was adopted by other methods. Szegedy \etal \cite{googlenet} proposed the GoogLeNet with a 22-layer structure and ``inception'' modules to replace the CNN in the R-CNN, which won the ILSVRC 2014 object detection task. Ouyang \etal \cite{ouyang2015deepid} proposed a deformation constrained pooling layer and a box pre-training strategy, which achieves an accuracy of $50.3\%$ on the ILSVRC 2014 test set. To accelerate the training of the R-CNN pipeline, Fast R-CNN \cite{girshick2015fast} is proposed, where each image patch is no longer wrapped to a fixed size before being fed into the CNN. Instead, the corresponding features are cropped from the output feature map of the last convolutional layer. In the Faster R-CNN pipeline \cite{girshick2015faster}, the bounding box proposals were generated by a Region Proposal Network (RPN), and the overall framework can thus be trained in an end-to-end manner. However, these pipelines are for object detection in still images. When these methods are applied to videos, they might miss some positive samples because the objects might not be of their best poses in each frame of the videos.

Object localization and co-localization~\cite{Prest:2012learning,Papazoglou:2013fast,Joulin:2014efficient,Kwak:2015unsupervised}, which have mainly focused on the YouTube Object Dataset (YTO) \cite{Prest:2012learning}, seems to be a similar topic to the VID task.
However, there are crucial differences between the two problems.
\textbf{1) Goal:} The (co)locolization problem assumes that each video contains only \emph{one} known (weakly supervised setting) or unknown (unsupervised setting) class and only requires localizing \emph{one} of the objects in each test frame. In VID, however, each video frame contains unknown numbers of objects instances and classes. The VID task is closer to real-world applications.
\textbf{2) Metrics:} Localization metric (CorLoc \cite{Deselaers:2010localizing}) is usually used for evaluation in (co)locolization, while mean average precision (mean AP) is used for evaluation on the VID task. The mean AP is more challenging to evaluate overall performances on different classes and thresholds.
% \textbf{3) Annotation:} YTO contains annotations on only several frames per video ($1$ frame per video on training set), while VID dataset contains annotations for different classes on every frame. The much richer annotation not only enables researches to investigate supervised deep learning methods which were not possible before, but also evaluates algorithm performances more precisely.
With these differences, the VID task is more difficult and closer to real-world scenarios. The previous works on object (co)localization in videos cannot be directly applied to VID.

There have also been methods on action localization. At each frame of human action video, the system is required to annotate a bounding box for the human action of interest. The methods that are based on action proposals are related to our work. Yu and Yuang \etal \cite{yu2015fast} proposed to generate action proposals by calculating actionness scores and solving a maximum set coverage problem. Jain \etal \cite{jain2014action} adopted the Selective Search strategy on super-voxels to generate tubulet proposals and proposed new features to differentiate human actions from background movements. In \cite{gkioxari2014finding}, candidate regions are fed into two CNNs to learn feature representations, which is followed by a SVM to make prediction on actions using appearance and motion cues. The regions are then linked across frames based on the action predictions and their spatial overlap.

Object tracking has been studied for decades \cite{possegger2014occlusion,li2015reliable,hong2015multi}. Recently, deep CNNs have been used for object tracking and achieved impressive tracking accuracy \cite{wang2015visual,nam2015learning,wang2016stct}. Wang \etal \cite{wang2015visual} proposed to create an object-specific tracker by online selecting the most influential features from an ImageNet pre-trained CNN, which outperforms state-of-the-art trackers by a large margin.
Nam \etal \cite{nam2015learning} trained a multi-domain CNN for learning generic representations for tracking objects. When tracking a new target, a new network is created by combining the shared layers in the pre-trained CNN with a new binary classification layer, which is online updated.
However, even for the CNN-based trackers, they might still drift in long-term tracking because they mostly utilize the object appearance information within the video without semantic understanding on its class.

% Attention models \cite{mnih2014recurrent,ba2014multiple,xu2015show} in deep learning have also drawn much attention in recent years. Although they can detect multiple objects in an image or a video sequentially, the temporal relations between the detections have not been studied. Therefore, they cannot be directly applied to the problem of object detection in videos.
% section related_works (end)

%%%%%%%%% BODY TEXT
\section{Method} % (fold)
\begin{figure*}[t]
    \centering
    \includegraphics[width=0.85\linewidth]{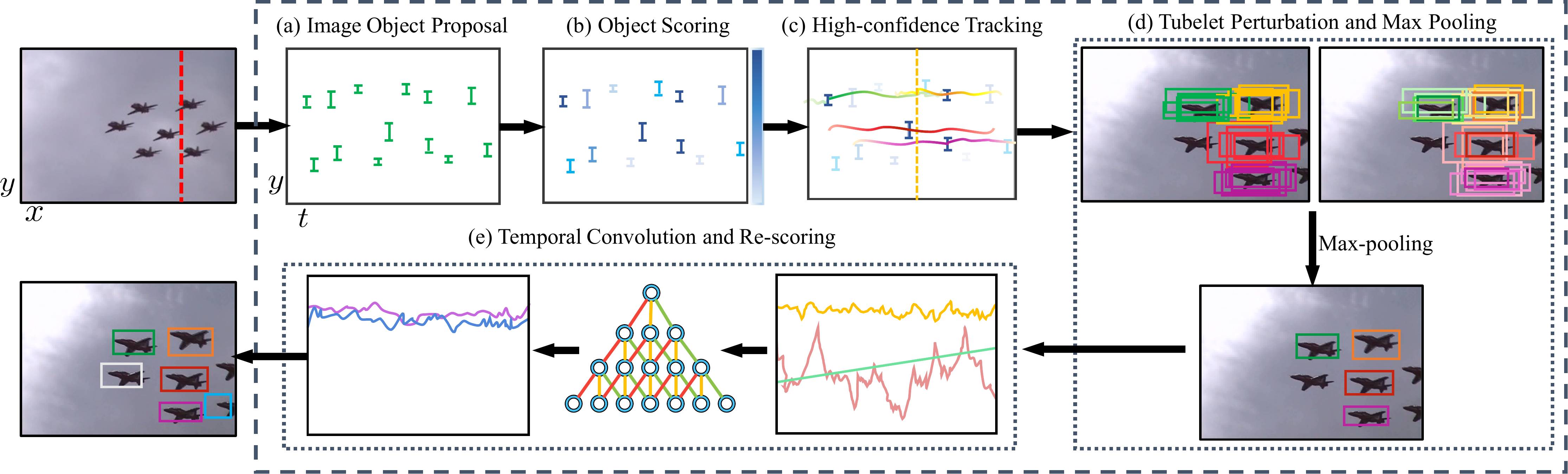}
    \caption{Video object detection framework. The proposed video object detection framework contains two major components. 1) The tubelet proposal component: (a), (b) and (c). 2) The tubelet classification and re-scoring component: (d) and (e). (a) In object proposal, class-independent proposals are generated on every frame. (b) An object scoring model outputs a detection score on every proposal. The darker the color, the higher the score. (c) High-confidence proposals are chosen as anchors for bidirectional tracking. (d) Tubelet boxes are perturbed by sampling boxes around them or replacing with original proposals. The spatial max-pooling on keeps the boxes with the highest detection scores for each tubelet box. (e) The time series of detection scores (Red), tracking score (Yellow) and anchor offsets (Green) are the inputs of the proposed temporal convolutional network. The purple line is the output of our network and blue line is the ground truth overlap value (supervision).}
    \label{fig:framework}
\end{figure*}

\label{sec:method}
In this section, we will introduce the task setting for object detection from video and give a detailed description of our framework design.
The general framework of video object detection system is shown in Figure~\ref{fig:framework}.
The framework has two main modules: 1) a spatio-temporal tubelet proposal module and 2) a tubelet classification and re-scoring module.
The two major components will be elaborated in Section~\ref{sub:tubelet_proposal} and Section~\ref{sub:tubelet_classification_and_rescoring}.

\subsection{Task setting} % (fold)
\label{sub:test_setting}
The ImageNet object detection from video (VID) task is similar to image object detection task (DET) in still images.
There are $30$ classes, which is a subset of $200$ classes of the DET task. All classes are fully labeled for each video clip.
For each video clip, algorithms need to produce a set of annotations $(f_i, c_i, s_i, b_i)$ of frame number $f_i$, class label $c_i$, confidence scores $s_i$ and bounding boxes $b_i$.
The evaluation protocol for the VID task is the same as DET task. Therefore, we use the conventional mean average precision (mean AP) on all classes as the evaluation metric.

\subsection{Spatio-temporal tubelet proposal} % (fold)
\label{sub:tubelet_proposal}
Objects in videos show temporal and spatial consistency.
The same object in adjacent frames has similar appearances and locations.
Using either existing object detection methods or object tracking methods alone cannot effectively solve the VID problem.
On one hand, a straightforward application of image object detectors is to treat videos as a collection of images and detect objects on each image individually.
This strategy focuses only on appearance similarities and ignores the temporal consistency. Thus the detection scores on consecutive frames usually have large fluctuations (Figure~\ref{fig:challenges} (a)).
On the other hand, generic object tracking methods track objects from a starting frame and usually on-line update detectors using samples from currently tracked bounding boxes.
The detectors in tracking methods mainly focus on samples within the video and usually tends to drift due to large object appearance changes (Figure~\ref{fig:challenges} (b)).

The spatio-temporal tubelet proposal module in our framework combines the still-image object detection and generic object tracking together. It has the discriminative ability from object detectors and the temporal consistency from object trackers.
The tubelet proposal module has $3$ major steps: 1) image object proposal, 2) object proposal scoring and 3) high-confidence object tracking.

\vspace{0.1cm}\noindent \textbf{Step 1. Image object proposal.}
The general object proposals are generated by the Selective Search (SS) algorithm\cite{uijlings2013selective}.
The SS method outputs around $2,000$ object proposals on each video frame. The majority object proposals are negative samples and may not contain objects. We use the ImageNet pre-trained AlexNet \cite{krizhevsky2012imagenet} model provided by R-CNN to remove easy negative object proposals where all detection scores of ImageNet detection $200$ classes are below a certain threshold.
In our experiments, we use $-1.1$ as threshold and around $6.1\%$ of the region proposals are kept, while the recall at this threshold is $80.49\%$. The image object proposal process is illustrated in Figure~\ref{fig:framework} (a).

\vspace{0.1cm}\noindent \textbf{Step 2. Object proposal scoring.}
Since the VID $30$ classes are a subset of DET $200$ classes, the detection models trained for the DET task can be used directly for VID classes.
Our detector is a GoogLeNet \cite{googlenet} pre-trained on ImageNet image classification data, and fine-tuned for the DET task. Similar to the R-CNN,  for each DET class, an SVM is trained with hard negative mining using the ``pool5'' features extracted from the model.
The $30$ SVM models corresponding to VID classes are used here to classify object proposals into background or one of the object classes. The higher the SVM score, the higher the confidence that the box contains an object of that class (Figure~\ref{fig:framework} (b)).

\vspace{0.1cm}\noindent \textbf{Step 3. High-confidence proposal tracking.}
For each object class, we track high-confidence detection proposals bidirectionally in the video clip.
The tracker we choose for this task is from \cite{wang2015visual}, which in our experiments shows more robust performance to object pose and scale changes.
The starting detections of tracking are called ``anchors'', which are chosen from the most confident box proposals from Step 2.
Starting from an anchor, we track backward to the first frame, and track forward to the last frame. Two tracklets are then concatenated to produce a complete track.
As the tracking moves away from the anchors, the tracker may drift to background and other objects, or may not keep up with the scale and pose changes of the target object. Therefore, we early stop the tracking when the tracking confidence is below a threshold (probability of $0.1$ in our experiments) to reduce false positive tracklets.
After getting a track, a new anchor is selected from the rest detections. Usually, high-confidence detections tend to cluster both spatially and temporally, therefore directly tracking the next most confident detection tends to result in tracklets with large mutual overlaps on the same object. To reduce the redundancy and cover as many objects as possible, we perform a suppression process similar to NMS. Detections from Step 2 that have overlaps with the existing tracks beyond a certain threshold (IOU $0.3$ in our experiment) will not be chosen as new anchors.
The tracking-suppression process performs iteratively until confidence values of all remaining detections are lower than a threshold (SVM score below $0$ in our setting).
For each video clip, such tracking process is performed for each of the $30$ VID classes.

With the above three major steps, we can obtain tracks starting from high-confidence anchors for each classes. The produced tracks are tubelet proposals for tubelet classification of later part of our framework.

% \begin{figure}[t]
%     \centering
%     % \includegraphics[]{}
%     \fbox{\rule{0pt}{2in} \rule{0.9\linewidth}{0pt}}
%     \caption{Detection and tracking weaknesses.}
%     \label{fig:det_track_weakness}
% \end{figure}

% subsection tubelet_proposal (end)
\subsection{Tubelet classification and rescoring} % (fold)
\label{sub:tubelet_classification_and_rescoring}
After the tubelet proposal module, for each class, we have tubelets with high-confidence anchor detections. A naive approach is to classify each bounding box on the tubelets using the same method as Step 2 before. In our experiment, this baseline approach has only modest performance compared to direct still-image object detection R-CNN. The reason for that is $4$-fold.

1) The overall number of bounding box proposals from tubelets is significantly smaller than those from Selective Search, which might miss some objects and result in lower recall on the test set.

2) The detector trained for object detection in still images is usually sensitive to small location changes (Figure~\ref{fig:framework} (d)) and a tracked boxes may not have a reasonable detection score even if it has large overlap with the object.
% In fact, even on ground truth bounding boxes, the detectors may not output robust detection scores.

3)  In the tracking process, we performed proposal suppression to reduce redundant tubelets. The tubelets are therefore more sparse compared than image proposals. This suppression may be conflict with conventional NMS. Because in conventional NMS, even a positive box has very low confidences, as long as other boxes with large mutual overlaps have higher confidence, it will be suppressed during NMS and will not affect the overall average precision. The consequence of early suppression is that some low-confidence positive boxes do not have overlaps with high confidence detections, thus are not suppressed and become false negatives.

4) The detection score along the tubelet has large variations even on ground truth tubelets (Figure~\ref{fig:challenges} (a)). The temporal information should be incorporated to obtain consistent detection scores.

To handle these problems in tubelet classification, we designed the following steps to augment proposals, increase spatial detection robustness and incorporate temporal consistency into the detection scores.

\vspace{0.1cm}\noindent \textbf{Step 4. Tubelet box perturbation and max-pooling.}
The tubelet box perturbation and max-pooling process is to replace tubelet boxs with boxes of higher confidence.
There are two kinds of perturbations in our framework.

The first method is to generate new boxes around each tubelet box on each frame by randomly perturbing the boundaries of the tubelet box. That is, we randomly sample coordinates for upper-left and bottom-right corners of a tubelet box. The random offsets for the corners are generated from two uniform distributions:
\begin{align}
    \Delta x &\sim U(-r\cdot w, r\cdot w),\\
    \Delta y &\sim U(-r\cdot h, r\cdot h),
\end{align}
where $U$ is uniform distribution, $w$ and $h$ are width and height of the tubelet box, and $r$ is the sampling ratio hyperparameter. Higher sampling ratio means less confidence on the original box, while lower sampling ratio means more confidence on the tubelet box. We evaluated performances of different sampling configurations (Section~\ref{sec:experiments}).

The second perturbation method is to replace each tubelet box with original object detections that have overlaps with the tubelet box beyond a threshold. This process is to simulate the conventional NMS process. If the tubelet box is positive box with a low detection score, this process can help bring back some positive boxes to suppress this box. The higher the overlap threshold, the more confidence on the tubelet box. We find this method really effective and different overlap threshold are evaluated in Section~\ref{sec:experiments}.

After the box perturbation step, all augmented boxes and the original tubelet boxes are scored using the same detector in Step 2.
For each tubelet box, only the augmented box with the maximum detection score is kept and used to replace the original tubelet box. The max-pooling process is to increase the spatial robustness of detector and utlize the original object detections around the tubelets.

\begin{figure}[t]
    \centering
    \includegraphics[width=0.8\linewidth]{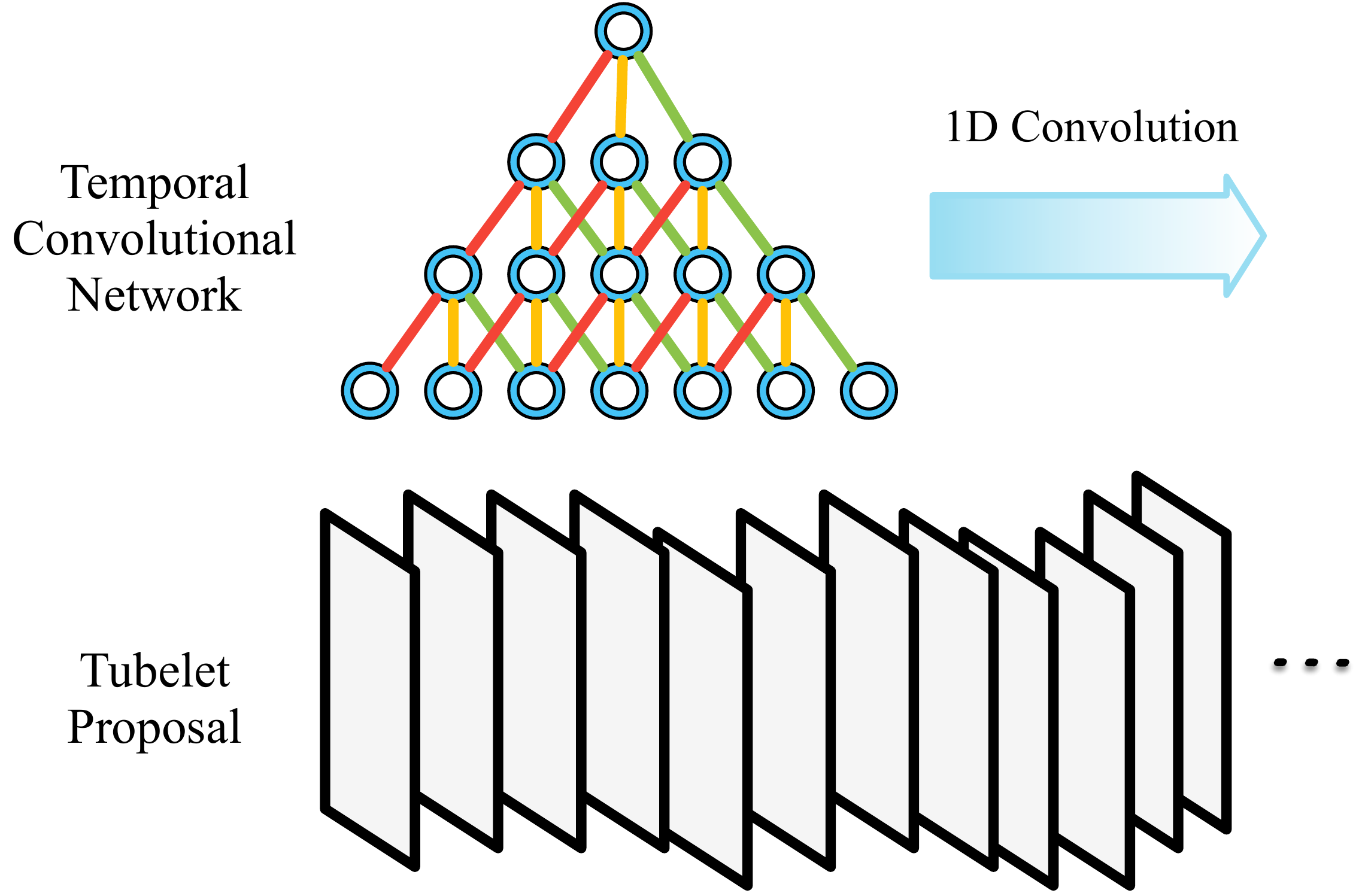}
    \caption{Temporal convolutional network (TCN). The TCN is a $1$-D convolutional network that operates on tubelet proposals. The inputs are time series including detection scores, tracking scores and anchor offsets. The output values are probablities that whether each tubelet box has overlap with ground truth above $0.5$.}
    \label{fig:tcn}
\end{figure}
\vspace{0.1cm}\noindent \textbf{Step 5. Temporal convolution and re-scoring.}
Even with the spatial max-pooling process, the detection scores along the same track might still have large variations. This naturally results in performance loss. For example, if tubelet boxes on ajacent frames all have high detection scores, it is very likely that the tublet box on this frame also has high confidence on the same object.
The still-image object detection framework does not take temporal consistency into consideration.

In our framework, we propose a Temporal Convolutional Network (TCN) that uses $1$-D serial features including detection scores, tracking scores, anchor offsets and generates temporally dense prediction on every tubelet box.

The structure of the proposed TCN is shown in Figure~\ref{fig:tcn}. It is a $4$-layer $1$-D fully convolution network that outputs temporally dense prediction scores on every tubelet box.
For each class, we train a class-specific TCN using the tubelet features as input.
The inputs are time series including detection scores, tracking scores and anchor offsets.
The output values are probablities whether each tubelet box contains objects of the class.
The supervision labels are $1$ if the overlap with ground truth is above $0.5$, and $0$ otherwise.

The temporal convolution learns to generate classification prediction based on the temporal features within the receptive field. The dense $1$-D labels provide richer supervision than single tubelet-level labels.
During testing, the continuous classification score instead of the binary decision values.
We found that this re-scoring process has consistent improvement on tubelet detection results.
% subsection tubelet_classification_and_rescoring (end)
% section method (end)

\section{Experiments} % (fold)
\label{sec:experiments}
\subsection{Dataset} % (fold)
\label{sub:dataset}
\noindent\vspace{0.1cm}\textbf{ImageNet VID.}
We utilize the ImageNet object detection from video (VID) task dataset to evaluate the overall pipeline and individual component of the proposed framework.
The framework is evaluated on the initial release of VID dataset, which consists of three distinct splits.
1) The training set contains $1952$ fully-labeled video snippets ranging from $6$ frames to $5213$ frames per snippet.
2) The validation set contains $281$ fully-labeled video snippets ranging from $11$ frames to $2898$ frame per snippet.
3) The test set contains $458$ snippets and the ground truth annotation are not publicly available yet.

We report all results on the validation set as a common convention for object detection task.

\noindent\vspace{0.1cm}\textbf{YTO dataset.}
In addition to the ImageNet VID dataset, we also evaluate our proposed framework on the YTO dataset for the object localization task.
The YTO dataset contains $10$ object classes, which are a subset of the ImageNet VID dataset.
Different from the VID dataset which contains full annotations on all video frames, the YTO dataset is weakly annotated, \ie each video is only ensured to contain one object of the corresponding class, and only very few frames are annotated for each video.
The weak annotation makes it infeasible to train the our models on the YTO dataset.
However, since the YTO classes are a subset of the VID dataset classes, we can directly apply the trained models on the YTO dataset for evaluation.

% subsection dataset (end)

\subsection{Parameter settings} % (fold)
\label{sub:parameter_settings}
\vspace{0.1cm}\noindent \textbf{Image object proposal.}
For image object proposal, we used the ``fast'' mode in Selective Search \cite{uijlings2013selective}, and resized all input images to width of $500$ pixels and mapped the generated box proposals back to original image coordinates.

The R-CNN provided AlexNet model is used to remove easy negative box proposals. We used threshold of $-1.1$ and remove boxes whose detections scores of all DET $200$ are below the threshold.
This process kept $6.1\%$ of all the proposals (about $96$ boxes per image).

\vspace{0.1cm}\noindent \textbf{Tracking.}
The early stopping tracking confidence is set to probability $0.1$. Therefore, if the tracking confidence is below $0.1$, the tracklet is terminated.
The minimum detection score for a new tracking anchor is $0$. If no more detection beyond this threshold, the whole tracking process ends for this class.
The track-detection suppression overlap is set to $0.3$.
For each video snippet, we chose at most $20$ anchors for tracking for each class.

\begin{figure}[t]
    \centering
    \includegraphics[width=0.9\linewidth]{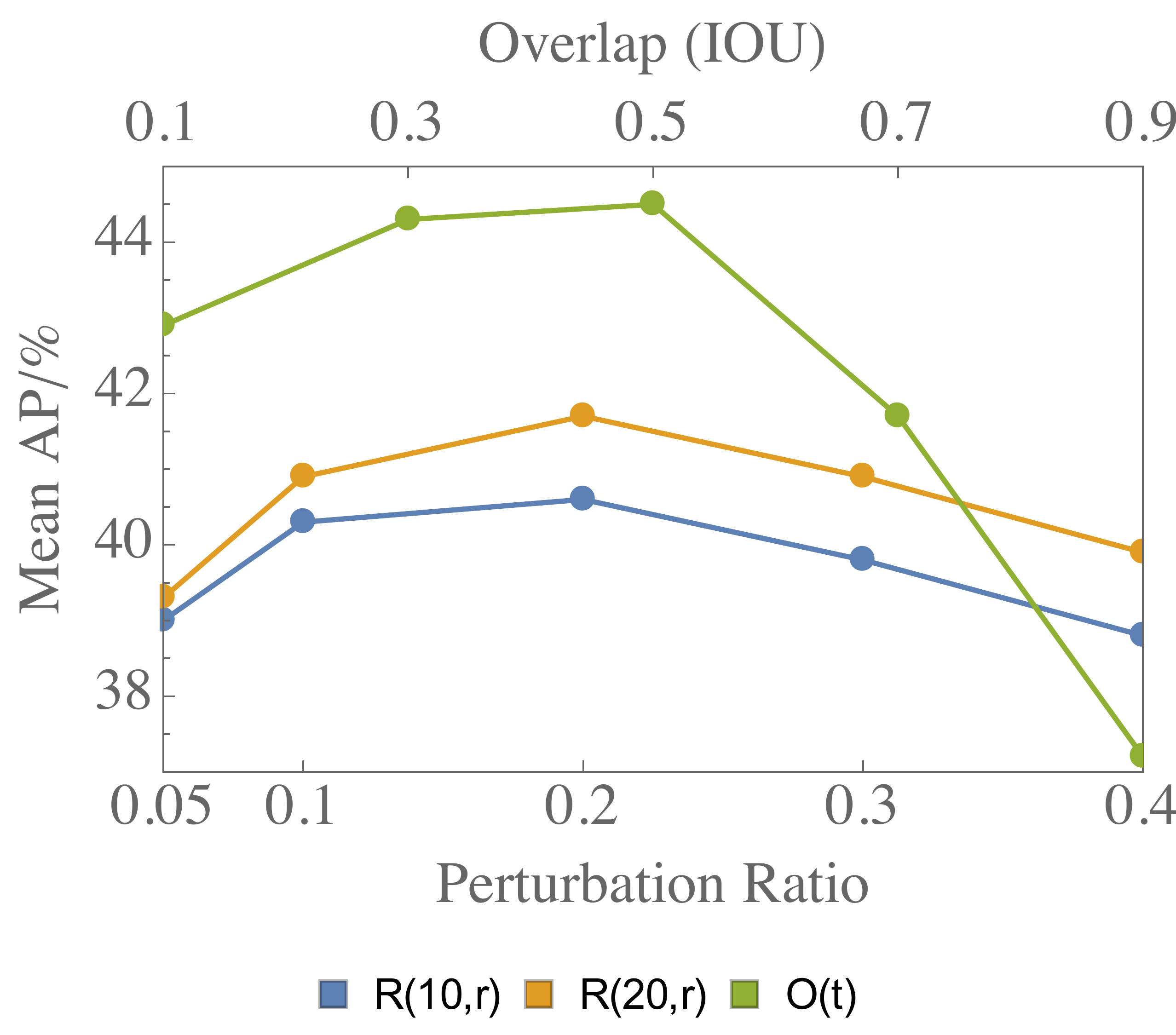}
    \caption{Performances of different max-pooling schemes. The blue and yellow lines are random sampling $10$ and $20$ samples per box with different perturbation ratios (bottom). The green line shows the performances of different overlap thresholds (top) for adding orginal proposals. Best viewed in color.}
    \label{fig:max-pooling_performance}
\end{figure}
\begin{table}[t]
\center
\footnotesize{
\begin{tabular}{M{1cm}M{1cm}|M{0.60cm}|M{0.60cm}|M{0.60cm}|M{0.60cm}|M{0.60cm}}

\hline
\multicolumn{2}{c}{\multirow{2}{*}{\textbf{mean AP/\%}}} & \multicolumn{5}{|c}{\textbf{Sampling ratio}} \\
\cline{3-7}
& & $0.05$ & $0.1$ & $0.2$ & $0.3$ & $0.4$ \\
\hline
\multirow{3}{*}{\textbf{\#samples}} & \multicolumn{1}{|c|}{Baseline} & \multicolumn{5}{c}{37.4} \\
\cline{2-7}
& \multicolumn{1}{|c|}{10} & 39.0 & 40.3 & 40.6 & 39.8 & 38.8 \\
\cline{2-7}
& \multicolumn{1}{|c|}{20} & 39.3 & 40.9 & \textbf{41.7} & 40.9 & 39.9 \\
\hline
\end{tabular}
}
\caption{Comparison of performances of different perturbation schemes.}
\label{tab:box_sampling}
\end{table}

\begin{table}[t]
\centering
\footnotesize{
\begin{tabular}{c|c|c|c|c|c|c}
\hline
\textbf{Overlap (IOU)} & 0.1 & 0.3 & 0.5 & 0.7 & 0.9 & Baseline\\
\hline
\textbf{mean AP/\%} & 42.9 & 44.3 & \textbf{44.5} & 41.7 & 37.2 & 37.4\\
\hline
\end{tabular}
}
\caption{Comparison of performances of different IOU threshold for adding original detections.}
\label{tab:iou_thres}
\end{table}

\vspace{0.1cm}\noindent \textbf{Tubelet perturbation.}
For tubelet box perturbation, we denote $R(n,r)$ for random perturbation with perturbation ratio $r$ and $n$ samples per tubelet box, and $O(t)$ for adding original proposals whose overlaps with the tubelet boxes beyond threshold $t$.

Different combinations of perturbation ratios and sampling numbers are evaluated as shown in Table~\ref{tab:box_sampling} and Figure~\ref{fig:max-pooling_performance}. $R(20,0.1)$ and $R(20,0.2)$ are chosen for later components.
For $O(t)$ schemes, $O(0.1)$ to $O(0.9)$ are evaluated (Figure~\ref{fig:max-pooling_performance} and Table~\ref{tab:iou_thres}). $O(0.5)$ is chosen for the framework.

\vspace{0.1cm}\noindent \textbf{Temporal convolutional network}
The TCN in our experiments has $4$ convolutional layers, the network structure is shown in Table~\ref{tab:tcn_param}.
The network initialization parameter and optimization parameter such as learning rate are manually adjusted on one class and remained unchanged for all $30$ classes.

The network raw detection score, tracking score and absolute anchor offsets (which is normalized by length of the track) are used as input feature for the TCN, without other preprocessing steps.

\begin{table}[t]
\centering
\scriptsize{
\begin{tabular}{p{1cm}|p{0.75cm}|p{0.75cm}|p{0.75cm}|p{0.9cm}|p{0.7cm}|p{0.7cm}}
\hline
\textbf{Layer} & Input & conv1 & conv2 & conv3 & conv4 & softmax \\
\hline
\textbf{Kernel size ($n\times t$)} & & $256\times5$ & $256\times5$ & $256\times7$ & $2\times3$ &  \\
\hline
\textbf{Output size ($c\times t$)} & $3\times50$ & $256\times50$ & $256\times50$ & $256\times50$ & $2\times50$ & $2\times50$\\
\hline
\end{tabular}
}
\caption{Temporal convolutional network (TCN) structure.}
\label{tab:tcn_param}
\end{table}
% subsection parameter_settings (end)
% section experiments (end)

\section{Results} % (fold)
\label{sec:results}

%%%%%%%%%%%%% TABLE MEAN AP %%%%%%%%%%%%%%%%%%
\begin{table*}[t]
\centering
\scriptsize{
\begin{tabular}{p{1.3cm}|p{0.5cm}p{0.5cm}p{0.5cm}p{0.5cm}p{0.5cm}p{0.5cm}p{0.5cm}p{0.5cm}p{0.5cm}p{0.5cm}p{0.5cm}p{0.5cm}p{0.5cm}p{0.5cm}p{0.5cm}p{0.5cm}}
\hline
Method & \rotatebox{60}{airplane} &  \rotatebox{60}{antelope} &  \rotatebox{60}{bear} &  \rotatebox{60}{bicycle} &  \rotatebox{60}{bird} &  \rotatebox{60}{bus} &  \rotatebox{60}{car} &  \rotatebox{60}{cattle} &  \rotatebox{60}{dog} &  \rotatebox{60}{domestic cat} &  \rotatebox{60}{elephant} &  \rotatebox{60}{fox} &  \rotatebox{60}{giant panda} &  \rotatebox{60}{hamster} &  \rotatebox{60}{horse} &  \rotatebox{60}{lion} \\
\hline
Still image & 64.50 &  71.40 &  42.60 &  36.40 &  18.80 &  62.40 &  \textbf{37.30} &  47.60 &  15.60 &  49.50 &  \textbf{66.90} &  66.30 &  58.20 &  74.10 &  25.50 &  29.00 \\
\hline
\hline
Baseline & 64.10 &  61.80 &  37.60 &  37.80 &  19.90 &  51.10 &  28.10 &  46.50 &  9.50 &  44.10 &  52.30 &  56.40 &  52.00 &  56.40 &  21.20 &  29.70 \\
s1: R(20,0.1) & 68.10 &  66.60 &  39.80 &  35.30 &  20.00 &  53.80 &  30.50 &  47.20 &  12.80 &  48.20 &  56.40 &  62.50 &  53.40 &  60.80 &  26.20 &  30.70 \\
s2: O(0.5) & 70.80 &  71.60 &  \textbf{43.20} &  37.00 &  20.40 &  \textbf{64.40} &  33.80 &  48.00 &  16.70 &  \textbf{54.50} &  62.70 &  70.80 &  57.10 &  73.30 &  26.00 &  30.70 \\
s3: s1 + s2 & 71.30 &  70.90 &  42.50 &  37.80 &  21.00 &  61.70 &  35.00 &  48.10 &  16.60 &  53.20 &  63.00 &  70.00 &  58.00 &  70.90 &  27.00 &  33.50 \\
TCN: s3 & \textbf{72.70} &  \textbf{75.50} &  42.20 &  \textbf{39.50} &  \textbf{25.00} &  64.10 &  36.30 &  \textbf{51.10} &  \textbf{24.40} &  48.60 &  65.60 &  \textbf{73.90} &  \textbf{61.70} &  \textbf{82.40} &  \textbf{30.80} &  \textbf{34.40} \\
\hline
\hline
Method & \rotatebox{60}{lizard} &  \rotatebox{60}{monkey} &  \rotatebox{60}{motorcycle} &  \rotatebox{60}{rabbit} &  \rotatebox{60}{red panda} &  \rotatebox{60}{sheep} &  \rotatebox{60}{snake} &  \rotatebox{60}{squirrel} &  \rotatebox{60}{tiger} &  \rotatebox{60}{train} &  \rotatebox{60}{turtle} &  \rotatebox{60}{watercraft} &  \rotatebox{60}{whale} &  \rotatebox{60}{zebra} &  \rotatebox{60}{mean AP} & \rotatebox{60}{\#win} \\
\hline
Still image & \textbf{68.70} &  1.90 &  50.80 &  34.20 &  \textbf{29.40} &  \textbf{59.00} &  \textbf{43.70} &  1.80 &  33.00 &  \textbf{56.60} &  66.10 &  61.10 &  24.10 &  64.20 &  45.30 & 7\\
\hline
\hline
Baseline & 38.70 &  1.90 &  41.40 &  34.10 &  19.60 &  45.50 &  10.90 &  1.30 &  12.80 &  48.10 &  64.90 &  52.40 &  17.20 &  63.60 &  37.40 & 0 \\
s1: R(20,0.1) & 51.80 &  1.50 &  44.60 &  33.30 &  15.10 &  51.40 &  26.20 &  1.80 &  18.70 &  49.00 &  65.80 &  57.10 &  22.20 &  67.40 &  40.60 & 0 \\
s2: O(0.5) & 62.80 &  \textbf{2.10} &  52.60 &  33.50 &  9.90 &  58.50 &  26.00 &  2.30 &  30.70 &  54.70 &  \textbf{67.70} &  62.50 &  23.60 &  66.30 &  44.50 & 5 \\
s3: s1 + s2 & 64.30 &  1.90 &  51.40 &  34.10 &  15.50 &  57.90 &  40.20 &  2.40 &  31.60 &  52.20 &  67.50 &  63.10 &  24.50 &  67.10 &  45.10 & 0 \\
TCN: s3 & 54.20 &  1.60 &  \textbf{61.00} &  \textbf{36.60} &  19.70 &  55.00 &  38.90 &  \textbf{2.60} &  \textbf{42.80} &  54.60 &  66.10 &  \textbf{69.20} &  \textbf{26.50} &  \textbf{68.60} &  \textbf{47.50} & \textbf{18}\\
\hline
\hline
\end{tabular}
}
\caption{Performances of different methods and experimental settings. ``s\#'' stands for different settings. $R(n,r)$ represents random sampling pertubation scheme with perturbation ration of $r$ and $n$ samples per tubelet box. $O(t)$ represents adding original proposals with overlap larger than threshold $r$.}
\label{tab:mean ap}
\end{table*}

\subsection{Quantitative Results on VID} % (fold)
\label{sub:quantitative_results}
\vspace{0.1cm}\noindent \textbf{Tubelet proposal and baseline performance}.
After obtaining the tubelet proposals, a straight-forward baseline approach for tubelet scoring is to directly evaluate tubelet boxes with the object detector for still images. The performance of this approach is $37.4\%$ in mean AP.
In comparison, the still-image object detection result is $45.3\%$ in mean AP.

\vspace{0.1cm}\noindent \textbf{Tubelet perturbation and spatial max-pooling}.
The performances of different tubelect box perturbation and max-pooling schemes are shown in Table~\ref{tab:box_sampling} and Table~\ref{tab:iou_thres}. From the tables, we can see that in most settings, both the random sampling and adding original proposals improves over the baseline approach.
Also, if the perturbation is too large or too small, the performance gain is small.
The reasons are: 1) large perturbation may result in replacing the tubelet box with a box too far away from it, and on the other hand, 2) small perturbation may obtain redundant boxes that reduce the effect of spatial max-pooling.

The best performance of random sampling scheme is $41.7\%$ of $R(20,0.2)$.
For replacing with original proposals, the best result is $44.5\%$ of $O(0.5)$.
It is worth noticing that the tubelet perturbation and max-pooling scheme does not increase the overall boxes of tubelet proposals but replaces original tubelet boxes with nearby ones with the highest confidences.

We also investigated the complementary properties of the two perturbation schemes.
The perturbed tubelets from the best settings of the both schemes ($41.7\%$ model from $R(20,0.2)$ and $44.5\%$ model from $O(0.5)$) are combined for evaluation.
The direct combination doubles the number of tubelets, and the performance increases from $41.7\%$ and $44.5\%$ to $45.2\%$, which is comparable to still-image object detection result with much fewer proposals on each image (around $1:38$).

\vspace{0.1cm}\noindent \textbf{Temporal convolution}.
Using the tubelets proposals, we trained a TCN for each of the $30$ classes for re-scoring. We use the continuous values of Sigmoid foreground scores as the confidence values for the tubelet boxes in evaluation.

For the baseline $37.4\%$ model, the performance increases to $39.4\%$ by $2\%$.
On the best single perturbation scheme proposal ($O(0.5)$), the performance increases from $44.5\%$ to $46.4\%$.
For combined tubelet proposals from two perturbation schemes, a $45.2\%$ model with $R(20,0.2)$ and $O(0.5)$ increases the performance to $47.4$, while a $45.1$ model with $R(20,0.1)$ and $O(0.5)$ increases to $47.5\%$.

From the results we can see that our temporal convolution network using detection scores, tracking scores and anchor offsets provides consistent performance improvement (around $2$ percents in mean AP) on the tubelet proposals.

Overall, the best performance on tubelet proposals by our proposed method is $47.5\%$, $2.2$ percents increase from still-image object detection framework with only $1/38$ the number of boxes for evaluation.
% subsection quantitative_results (end)
\begin{figure*}[t]
    \centering
    \includegraphics[width=0.9\linewidth]{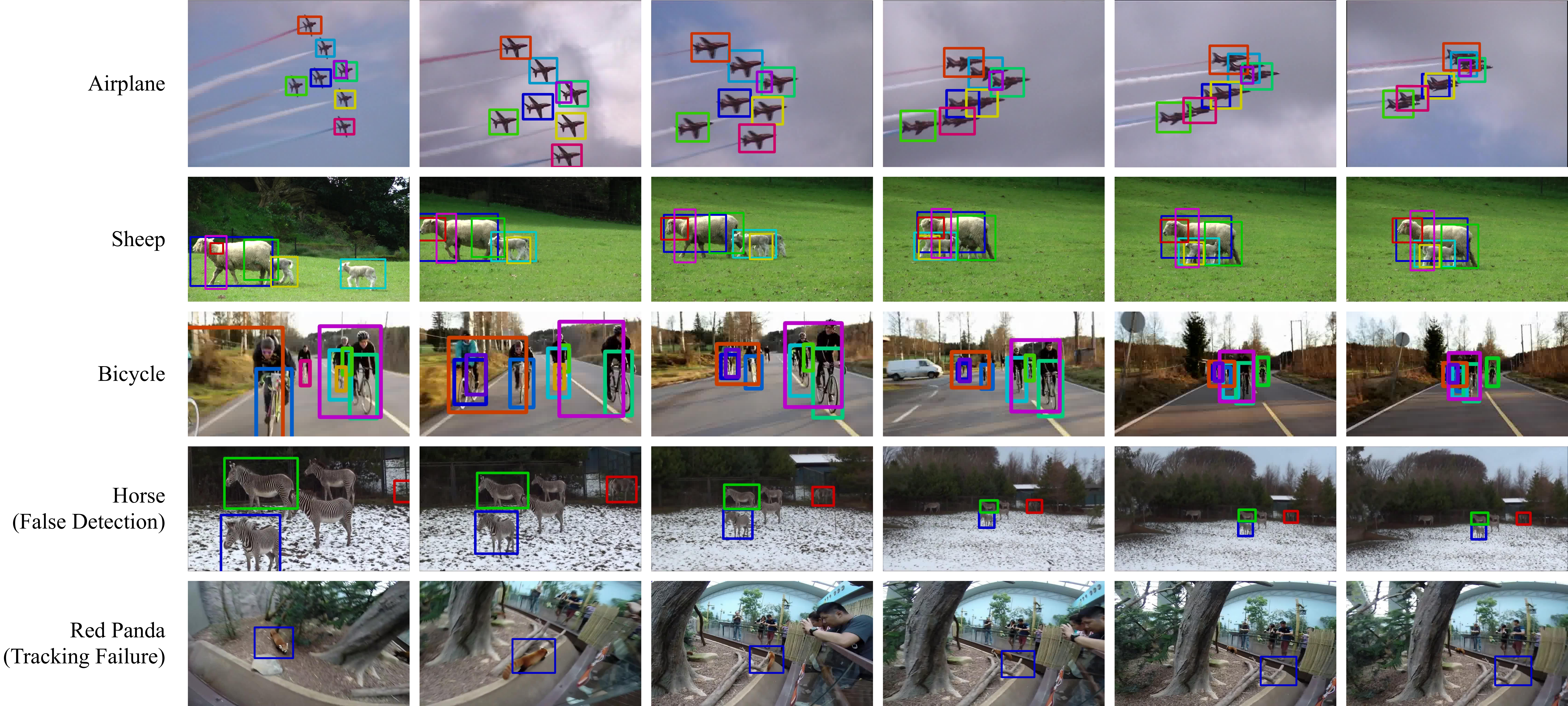}
    \caption{Qualitative results of tubelet proposals. The first three rows are positive bounding boxes of tubelet proposals generated by our framework. The proposal boxes usually aggregate on ground truth objects while keep sparsely allocated to cover as many objects as possible. The last two rows shows some failure cases of tubelet proposals. The first kind of failure cases are false detections (for example, mis-detect zebras as horses). The second kind of failure cases are tracking failures. Tracker drifts to background objects due to large scale changes of target objects while the mis-tracked targets are not confident enough to start new tracking processes.}
    \label{fig:tubelet_proposal}
\end{figure*}

\subsection{Qualitative Results on VID} % (fold)
\label{sub:qualitative_results}
\vspace{0.1cm}\noindent \textbf{Tubelet proposal}.
The tubelet proposal results are shown in Figure~\ref{fig:tubelet_proposal}.
Figure~\ref{fig:tubelet_proposal} (a) shows the positive tubelets obtained from tubelet proposal module, and Figure~\ref{fig:tubelet_proposal} (b) shows the negative samples.

From the figure we can see, positive samples usually aggregate around objects while still appear sparse compared to dense proposals from Selective Search.
The sparsity comes from the track-proposal suppression process performed in tracking to ensure the tubelet covers as many objects as possible.

With the frame index increases, some tracks will disappear while others may be added, which results from the early stopping for low tracking confidence and new anchor selections.

As for the negative samples, the tubelet are much fewer (in fact, some videos do not have tubelet proposals for some classes) and isolated. This is because we only start tracking on high-confident anchors. This largely reduces the number of false positives and significantly save inference time in later steps in the framework.

\begin{figure}[h]
    \centering
    \includegraphics[width=\linewidth]{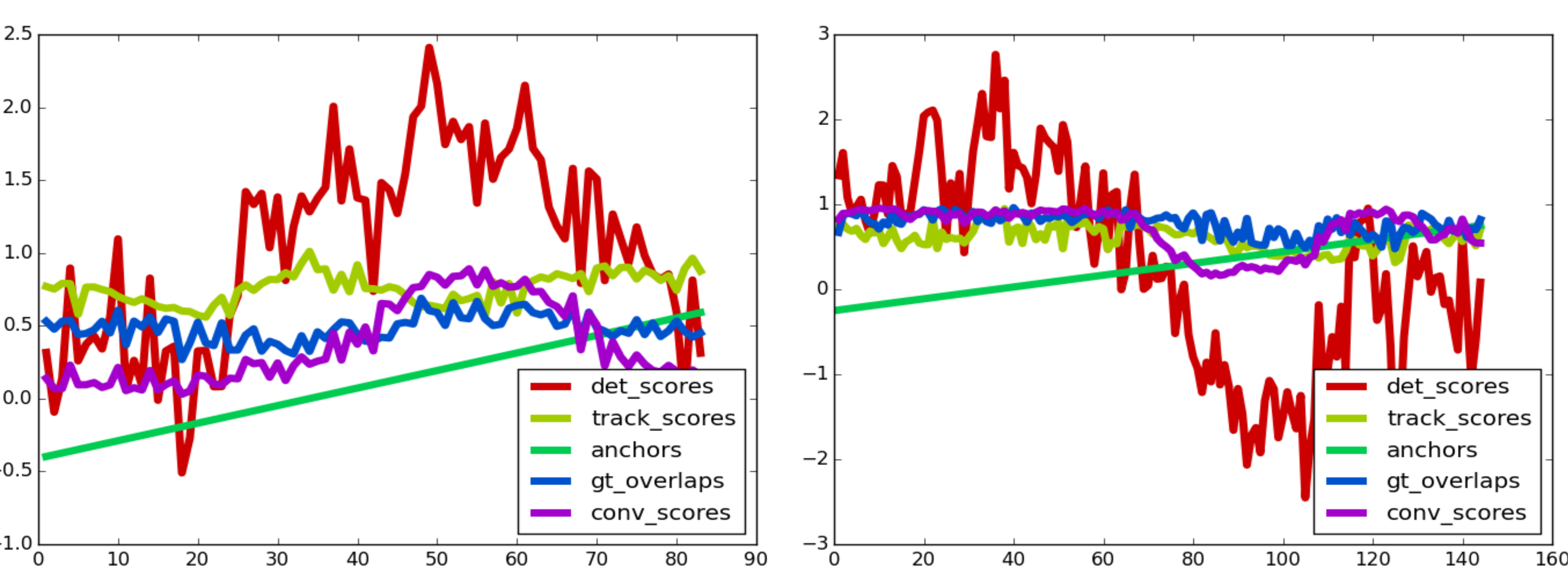}
    \caption{Qualitative results of temporal convolution. The time series of detection scores, tracking scores and absolute values of anchor offsets are the inputs for TCN. The blue line are overlaps with ground truth annotations and purple lines are the output of TCN. The detection scores have large temporal variations while the TCN output has temporal consistency and comply better to ground truth overlaps.}
    \label{fig:temp_conv}
\end{figure}

\vspace{0.1cm}\noindent \textbf{Temporal convolution}.
In Figure~\ref{fig:temp_conv}, we show some examples of results of temporal convolution. Each plot shows the tubelet scores (detection score, tracking score and anchor offsets) and the output probability scores of the TCN network.

The detection score shows significant variations across frames. A tubelet box may have significantly low detection score even if adjacent frames have high detection values.
However, after temporal convolution, the output probability curve are much more temporally consistent.
Compare to detection scores, the probability output of our network conforms better with the ground truth overlaps, which shows the effectiveness of our re-scoring module.
% subsection qualitative_results (end)
% section results (end)

\subsection{Evaluation on YouTube-Objects (YTO) dataset} % (fold)
\label{sub:eval_yto}
In order to show the effectiveness of our proposed framework, we applied the models trained on the VID task directly on the YTO dataset and compared with the state-of-the-art works in Table~\ref{tab:yto_scores}. The localization metric CorLoc \cite{Deselaers:2010localizing} is used for evaluation as a convention on YTO.

\begin{table}[h]
    \caption{Localization performances on the YTO dataset}
    \label{tab:yto_scores}
    \centering
    \tiny
    \begin{tabular}{l|p{0.15cm}p{0.15cm}p{0.15cm}p{0.15cm}p{0.15cm}p{0.15cm}p{0.15cm}p{0.15cm}p{0.15cm}p{0.15cm}|c}
    \hline Method & aero & bird & boat & car & cat & cow & dog & horse & mbike & train & Avg. \\
    \hline
    \hline
    Prest \etal \cite{Prest:2012learning} & 51.7 & 17.5 & 34.4 & 34.7 & 22.3 & 17.9 & 13.5 & 26.7 & 41.2 & 25.0 & 28.5 \\
    Joulin \etal \cite{Joulin:2014efficient} & 25.1 & 31.2 & 27.8 & 38.5 & 41.2 & 28.4 & 33.9 & 35.6 & 23.1 & 25.0 & 31.0 \\
    Kwak \etal \cite{Kwak:2015unsupervised} & 56.5 & 66.4 & 58.0 & 76.8 & 39.9 & \textbf{69.3} & 50.4 & 56.3 & 53.0 & 31.0 & 55.7 \\
    \hline
    Baseline & 92.4 & 68.4 & 85.4 & 75.8 & \textbf{77.3} & 18.6 & 87.2 & 87.3 & 84.2 & 72.8 & 74.9 \\
    Ours (TCN:s3) & \textbf{94.1} & \textbf{69.7} & \textbf{88.2} & \textbf{79.3} & 76.6 & 18.6 & \textbf{89.6} & \textbf{89.0} & \textbf{87.3} & \textbf{75.3} & \textbf{76.8} \\
    \hline
    \end{tabular}
\end{table}

From the table, we can see that our proposed framework outperforms by a large margin.
This is because the ImageNet datasets (CLS, DET and VID) provide rich supervision for feature learning, and the trained networks have good generalization capability on other datasets.

The full framework has around $2\%$ improvement over the baseline approach on the YTO dataset, which is consistent with the results on VID.
% subsection eval_yto (end)

\section{Conclusion} % (fold)
\label{sec:conclusion}
In this work, we propose a complete multi-stage pipeline for object detection in videos.
The framework efficiently combines still-image object detection with generic object tracking for tubelet proposal.
Their relationship and contributions are extensively investigated and evaluated.
Based on tubelet proposals, different perturbation and scoring schemes are evaluated and analyzed.
A novel temporal convolutional network is proposed to incorporate temporal consistency and shows consistent performance improvement over still-image detections.
Based on this work, a more advanced tubelet-based framework is further developed which won the ILSVRC2015 ImageNet VID challenge with provided data \cite{kang2016tcnn}.

% section conclusion (end)

\section*{Acknowledgement} % (fold)
\label{sec:acknowledgement}
This work is partially supported by SenseTime Group Limited, and the General Research Fund sponsored by the Research Grants Council of Hong Kong (Project Nos. CUHK14206114, CUHK14205615, CUHK14203015, CUHK14207814).
% section acknowledgement (end)

{\small
\bibliographystyle{ieee}
\bibliography{egbib}
}

\end{document}